# A Comparative Study of Feature Selection Methods for Dialectal Arabic Sentiment Classification Using Support Vector Machine

**Omar Al-Harbi1**[†]

oalhabi@Jazanu.edu.sa

*Jazan University* [†]

**Summary**

Unlike other languages, the Arabic language has a morphological complexity which makes the Arabic sentiment analysis is a challenging task. Moreover, the presence of the dialects in the Arabic texts have made the sentiment analysis task is more challenging, due to the absence of specific rules that govern the writing or speaking system. Generally, one of the problems of sentiment analysis is the high dimensionality of the feature vector. To resolve this problem, many feature selection methods have been proposed. In contrast to the dialectal Arabic language, these selection methods have been investigated widely for the English language. This work investigated the effect of feature selection methods and their combinations on dialectal Arabic sentiment classification. The feature selection methods are Information Gain (IG), Correlation, Support Vector Machine (SVM), Gini Index (GI), and Chi-Square. A number of experiments were carried out on dialectical Jordanian reviews with using an SVM classifier. Furthermore, the effect of different term weighting schemes, stemmers, stop words removal, and feature models on the performance were investigated. The experimental results showed that the best performance of the SVM classifier was obtained after the SVM and correlation feature selection methods had been combined with the uni-gram model.

## 1. Introduction

Sentiment analysis or opinion mining field is a task of Natural Language Processing (NLP), which concerns with detection of subjectivity in textual information [1]. Sentiment analysis attempts to classify the texts into subjective or objective classes and detect the positive or negative opinions in the subjective texts. Research on sentiment analysis of English has achieved considerable progress, whereas it is still limited in the Arabic language. The Arabic language is morphologically complex, and it may contain different distinct dialects [2]. El-Beltagy and Ali [3] highlighted many significant issues of sentiment analysis in Arabic languages such as the existence of dialects, the lack of Arabic dialects resources and tools, the limitation of Arabic sentiment lexicons, using compound phrases and idioms, etc.

Machine learning approach or supervised approach uses statistics to learn from labelled instances of texts or sentences [4]. Machine learning approach includes several algorithms that have been applied to sentiment analysis field such as Support Vector Machine (SVM), Naïve Bayes (NB), Decision Trees, K-Nearest Neighbor (K-NN), Maximum Entropy, etc. In the sentiment analysis literature, SVM has been reported as the best learning algorithm [5, 6]. The Hybrid approach combines both machine learning and semantic orientation. In this approach, the machine learning techniques are used as a classifier along with semantic orientation features to conduct the training model. The hybrid approach has been widely employed in the literature of Arabic language such as [7-11].

In this work, the machine learning approach to deal with dialectal Arabic sentiment analysis was used. The first step to use any machine learning algorithm in sentiment analysis is converting the documents to a feature vector. Features can be useful only if they are fast to compute and preserve useful discriminatory information [12]. In sentiment analysis, there are different feature models have been used; however, it was noted that N-gram model is the most used with machine learning algorithms [5, 13-15].

However, one of the challenging issues that face the machine learning algorithms in sentiment analysis and more generally in text classification is the huge number of the created features. A feature can positively or negatively affect the performance of classification based on its relevancy and redundancy with respect to the class labels [16-19]. Therefore, feature selection methods are required to make the text classifiers efficient and more accurate by selecting the most relevant and discriminating feature vector [16]. Feature selection methods identify the most informative features based on measuring the goodness of a feature subset with which the best performance can be obtained. Liu and Motoda in [20] defined features selection as a process that chooses an optimal subset of features according to a certain criterion. The objective of using feature selection is to improve classification performance (speed, learning, and accuracy), and a better understanding of the underlying process that generated the data [12].

In general, Many feature selection methods have been proposed and applied to sentiment analysis, i.e., Document frequency (DF), Information Gain (IG), Fisher Score, Mutual Information (MI), Relief-F, Chi-Square, Principal Components Analysis, SVM, etc. Regarding to English sentiment classification, different automated and manual techniques have been proposed and used to select the optimal feature sets. Whereas, based on a thorough review of previous studies on Arabic sentiment analysis, there has been limited work on feature selection, especially for dialectal Arabic. Furthermore, several powerful techniques and their combinations have not been investigated yet.

This work investigated the effect of several feature selection methods and their combination on the performance of an SVM classifier which was trained on dialectal Arabic reviews. The key idea behind combining feature selection methods is to take advantage of each method's capability to improve feature selection performance [20]. Each method has its weaknesses and strengths that affect the feature selection accuracy. Thus, a hybrid approach that can benefit from the strengths and avoid weaknesses of the selection methods was used. Furthermore, we investigated the effect of different term weighting schemes, stemmers, stop words removal, and feature models on the performance of the classifier.

The paper is conducted as follows. Section 2 presents related work. Section 3 introduces the methodology and experiment setting. Section 4 discusses experimentations and results. Finally, Section 5 presents the conclusions of this work.

## 2. Related Work

In the literature of sentiment analysis, various machine learning approaches have been applied [1, 18, 21-24]. Feature selection task involved in machine learning classification as a method to improve the performance by reducing the dimensionality. In the English language, there exist a considerable amount of published work have used different methods of feature selection methods to improve the performance of sentiment analysis [13, 25-28]. In contrast, little attention has been given to feature selection effect on dialectal Arabic sentiment analysis.

Abbasi et al. [7] developed a hybridized genetic algorithm that incorporates the IG heuristic for feature selection called Entropy-Weighted Genetic Algorithm (EWGA). Their method was designed to evaluate different feature sets consisting of syntactic and stylistic features for English and Arabic dataset. They applied their proposed methods and features to a multi-language web forum at the document level. They reported that using EWGA with SVM obtained high performance levels, with accuracies of over 93% for the Arabic language.

Duwairi and El-Orfaili [29] investigated the effect of feature correlation on Arabic sentiment classification performance. Different N-gram models of words and characters were used for text representation. Both MSA and dialect were present in the dataset that includes political and movie reviews. Three classifiers were used to classify the reviews, namely, SVM, Naïve Bayes, and K-NN. Significant improvement has been obtained with applying the top 1200 correlated features and word N-gram using Naïve Bayes classifier. SVM and Naïve Bayes classifiers showed better performance, where Naïve Bayes classifier yielded the highest accuracy with 97.2%.

Shoukry and Rafea [9] proposed a hybrid approach for sentiment classification of Egyptian dialect. The feature vector was built of different N-gram models and polarity scores. For reducing the features, they assigned a frequency threshold for each feature through 20,000 tweets. They used an SVM classifier to train the data. The results showed that the best performance is 84%, and obtained with a combination of uni-gram, bi-gram, and tri-gram model.

Khalil et al. [30] introduced in their work an experimental study to examine the effect of different text representation schemes on machine learning methods for Arabic sentiment analysis. They experimented with various datasets which contain tweets written in MSA, Egyptian dialect, and Saudi dialect. Three classifiers were chosen, namely, SVM, compliment NB, and multinomial NB. For selecting the optimal features, they employed IG method and other basic data preprocessing techniques. However, IG did not accomplish a significant improvement in performance. The result showed that a combination of uni-gram and bi-gram obtained the highest accuracy with the IDF weighting scheme.

Aliane et al. [31] investigated the importance of using a genetic algorithm feature selection approach in Arabic sentiment analysis. They used LABR dataset which contains MSA and dialectal reviews about books. The experiment was performed with five Machine Learning supervised algorithms SVM, Naive Bayes (NB), Multinomial Naive Bayes (MNB), Stochastic Gradient descent and Decision Trees. As noted from the results, the accuracy improved after applying the genetic algorithm, where MNB yielded the highest accuracy (94%) with uni-grams and TF-IDF weighting scheme.

El-Naggar et al. [32] presented a hybrid approach for sentiment analysis of MSA and Egyptian Dialectal tweets. They used IG method to select the relevant features which have been fed to SVM and Random Forest classifiers. As well as, various sentiment features were used based on different lexicons. Based on their results, the approach obtained 90% accuracy with superiority to [33] and [34].

Omar et al. [35] presented an empirical comparison of seven feature selection methods (IG, Principal Components Analysis, Relief-F, Gini Index, Uncertainty, Chi-squared, and Support Vector Machines (SVMs)) for Arabic sentiment analysis. They used three classifiers (SVM, Naive Bayes, and K-nearest neighbor) for MSA and dialectal reviews about movies. Authors observed that a significant improvement on the performance when feature selection methods are used. Their experimental results showed that SVM classifier with SVM-based feature selection method obtained the highest accuracy with 92.4%.

Based on the review of many previous studies on Arabic sentiment analysis, it can be noticed that most of the studies have focused only on the comparison between different classifiers on MSA texts. Furthermore, a few studies have addressed the feature selection methods to resolve the problem of the dimensionality in dialectal Arabic. Additionally, the feature selection methods have been individually utilized and investigated, without considering the idea of combining these methods. The combination can exploit the advantages of the feature selection methods and avoid their disadvantages [20]. Therefore, besides investigating the effect of different individual methods of feature selection, this work also examined their combinations on the dialectal Arabic sentiment analysis. Moreover, the effect of different weighting schemes, stemmers, and feature models on the performance was investigated.

## 3. Methodology

This section describes the methodology used to classify sentimental Arabic reviews. In this work, the SVM algorithm was used to classify the reviews into positive or negative classes. The work also used a dataset presented by [11] for experimenting. The dataset was pre-processed to perform the other processes efficiently. Then, to find the best text representation, we explored the effect of using different stemming methods, weighting schemes, and stop word removal. Subsequently, the reviews were turned into a feature model to be trained and tested. After that, different feature selection methods and their combinations were applied to select the optimal feature subset. For experimenting, we used the Rapidminer software [36], which is a software platform that includes a valuable set of machine learning algorithms and tools for data and text mining. The following subsections discuss in details the functionality of each component in the methodology:

### 3.1 Dataset

To train the classifier, we need an annotated dataset. In our case, a publicly available dataset for Jordanian dialect[1] was used [11]. The dataset is annotated on the document level, and it considers only two polarity classes, which are positive and negative. To balance the dataset we randomly selected 2400 reviews of which 1200 were positive, and 1200 were negative. The data consists of MSA and colloquial Jordanian reviews about various domains (restaurants, shopping, fashion, education, entertainment, hotels, motors, and tourism).

---
[1] The dataset is accessible online:
https://bit.ly/2FEGs1B

### 3.2 Pre-Processing

The pre-processing stage included removing noise from data, normalization, and tokenization. The process of removing noise from data includes removing misspellings, repeated letters, diacritics, punctuations, numerals, English words, and elongation. After that, a normalization process was applied to particular letters, for example the letters ( أ, آ, إ) were converted to (ا), the letters (ى, ئ) were converted to (ي), the letter (ة) was converted to (ه), and finally the letter (ؤ) was converted to (و). Tokenization is the process of dividing a given text into a set of words (tokens) which are separated by spaces. To find the best text representation, this work investigated three term-weighting schemes, namely, Term Frequency (TF), Term Frequency-Inverse Document Frequency (TF-IDF), and Binary Term Presence (BTP). Furthermore, we evaluated the effect of stop words removal and stemming (light stemming and root stemming) on the performance of the classifier. For the experimenting, list of stop words and the stemmers provided with Rapidminer were used.

### 3.3 Features Representation

To use a machine learning classifier to resolve the problem of the sentiment analysis, we need a suitable text representation model. This model is often called a vector model or feature model, which is represented by a matrix of term weights. Several feature models have been proposed in the literature such as semantic, stylistic, and syntactic [7]. This work used what so-called basic features, namely, N-grams (of different n words lengths). This model has been widely investigated in the literature of sentiment analysis and showed that it can improve the performance such as in the work of [5, 37]. This work experimented with uni-grams, bi-grams, and a combination of uni-gram and bi-gram. The values of n in uni-gram and bi-gram were one and two words respectively.

### 3.4 Feature Selection Methods

The feature vector plays a significant role in the performance of the classification. Notably, that features if, by including or excluding them, the performance would improve or degrade. The relevant feature is essential to the process of training since it has an informative aspect that would improve the classification. Whereas, the irrelevant ones are less informative, so including them might negatively affect the performance. Deciding which of the features relevant or irrelevant is the task of the feature selection methods. This work investigated five selection methods, namely, IG, correlation, Chi-Square, Gini Index, and SVM. For experimenting, the top n features created by the selection methods were selected. The value of n was specified based on two scenarios; namely, number of features less than the instances, and number of features greater than the instances. We also experimented with different combinations of these methods to take advantage of each method's capability. In this case, different selection methods were combined in a sequence mode to select the most relevant features. More details about the feature selection methods are presented in the following subsections.

### 3.4.1 Information Gain (IG)

IG is an important feature selection method which measures how much the feature is informative about the class. IG represent the uncertainty reduction in identifying category by knowing when the value of the feature. IG is a ranking score method which can be calculated for a term by Equation 1:

$$IG(w) = -\sum_{j=1}^{k} P(C_j)\log(P(C_j)) + P(w)\sum_{j=1}^{k} P(C_j|w)\log(P(C_j|w)) + P(\overline{w})\sum_{j=1}^{k} P(C_j|\overline{w})\log(P(C_j|\overline{w})) \quad (1)$$

Where $C$ is a class attribute and can be $K$ classes, it is denoted as $\{C_1, ..., C_k\}$. The probability of a feature $P(C_j)$ is the fraction of number of documents that belongs to class $C_j$ out of total documents, and $P(w)$ is the fraction of documents in which word $w$ occurs. $P(C_j|w)$ is computed as the fraction of documents from class $C_j$ that has word $w$. Where $\overline{w}$ means that a document does not contain the word $w$.

### 3.4.2 Correlation

Correlation is a statistical method to calculate the relevance of the feature with respect to the class feature. The correlation coefficient takes a value between -1 and +1 that represent the strength of association between two features. The positive correlation means as one variable gets larger the other gets larger. Whereas, the negative correlation means as one gets larger the other gets smaller. If the correlation is zero, it means there is no association between the features. Suppose we have two features observations $\{x_1, ..., x_n\}$ and $\{y_1, ..., y_n\}$ then that the correlation coefficient $r$ is calculated as in Equation 2.

$$r = r_{xy} = \frac{\sum x_i y_i - n\bar{x}\bar{y}}{(n-1)s_x s_y} \quad (2)$$

Where, $n$ is the sample size, $\bar{x}$ and $\bar{y}$ are the means of $x$ and $y$ respectively. $s_x$ and $s_y$ are the standard deviations of $x$ and $y$.

### 3.4.3 Chi-Square

Chi-Square is a statistical technique used to measure the difference between the expected frequencies and the observed frequencies for two events. In feature selection, the two events are occurrence of the term and occurrence of the class. In Arabic sentiment analysis, a few studied investigated the effect of using Chi-Square feature selection method such as. The value for each term $t$ with respect to the value of class $c$ is calculated by the Equation 3:

$$X^2(t,c) = \frac{N(AD - BC)^2}{(A+C)(B+C)(A+B)(C+D)} \quad (3)$$

Where, $N$ is the total number of documents, $A$ is the number of $t$ occurrences and $c$ occurrences, $B$ is the number of $t$ occurrences without $c$, $C$ is the number of $c$ occurrences without $t$, $D$ is the number of non-occurrences of $t$ and $c$.

### 3.4.4 Gini Index

Gini index is a feature selection method which measures the purity of the features with respect to the class [38]. The purity refers to the discrimination level of a feature to distinguish between the possible classes [39]. This feature selection method measures the purity when using a chosen feature. For a feature $t$, the Gini index is calculated by the Equation 4:

$$GI(t_i) = \sum_{j=1}^{m} p(t_i|C_j)^2 p(C_j|t_i)^2 \quad (4)$$

Where, $m$ is the number of classes, $p(t_i|C_j)$ is the term $t_i$ probability given class $c_j$, $p(C_j|t_i)$ is the class $c_j$ probability given the term $t_i$.

### 3.4.5 SVM-based Feature selection

Intuitively, using a feature selection method prior to a classifier which both use same prediction model is an attractive approach [40]. Thus, we assume using the SVM-based feature selection method along with an SVM classifier will obtain better performance. SVM method calculates the relevance of the features by computing the coefficients of a hyperplane as feature weights. The features are represented by vector $\vec{x}$ which contain a number of distinct features with respect to the class feature $y_i \in \{+1, -1\}$. The hyperplane is used to separate the two classes, and $w$ is the normal vector to the hyperplane. The SVM with input $\vec{x}$ and $\vec{w}$ gives the output as in Equation 5:

$$\vec{w}.\vec{x} - b = 0 \quad (5)$$

The margin is the distance between two parallel hyperplanes that separate the two classes of data. The margin m is calculated as in Equation 6:

$$m = \frac{1}{2}\|\vec{w}\|^2 \quad (6)$$

The distance between two hyperplanes must be as large as possible, so to maximize the margin we need to minimize $\|\vec{w}\|$. To do so we add the following constraints:

$$\vec{w}.\vec{x_i} - b \geq 1, if\ y_i = 1 \quad (7)$$

$$\vec{w}.\vec{x_i} - b \leq 1, if\ y_i = -1 \quad (8)$$

To find $\vec{w}$, the Lagrangian formulation is used. Then, the value of $\vec{w}$ will be as in Equation 9:

$$\vec{w} = \sum_{i=1}^{n} \alpha_i y_i x_i \quad (9)$$

The resulting $\vec{w}$ represents the weight vector of the features. Then, only the features with weights that satisfy a threshold will be selected.

### 3.5 Sentiment Classification

The goal of the classification is to categorize input data into predefined classes. This work concerns only with two classes; they are positive and negative. Next step after transforming the data into feature space is selecting the suitable learning classifier. In this work, the SVM classifier was used as it is one of the most well-known classifiers in recent years. In sentiment analysis, the SVM classifier has outperformed other classifiers in many researches such as in [5, 7, 41]. The SVM is a machine learning technique for binary classification problems introduced by [42]. In general, SVM has the advantage of overfitting protection, and its capability to handle large feature spaces [43]. In the experiments, an integrated software for linear kernel SVM called LIBSVM was used [44].

## 4. Experiments and Results

### 4.1 Experiment Design

Different experiments were undertaken to examine the effect of feature selection methods and some pre-processing techniques on the dialectal Arabic sentiment analysis. As mentioned earlier, the dataset include reviews which were annotated on the document level, and consist of 2400 reviews of which 1200 were positive, and 1200 were negative. The experiments were implemented using the SVM classifier, and we had employed the linear kernel as it empirically gave the best performance. The experiments included five stages of classification in which different weighting schemes, text representations, and feature selection methods were investigated. First, this work investigated the three weighting schemes, namely, TF, TF-IDF, and BTP. Second, the effect of pre-processing techniques on the performance was evaluated. These techniques include stop words removal and two methods of stemming (light stemming and root stemming) which are available in Rapidminer. Third, we investigated three text representation using N-gram models; they are uni-gram, bi-gram, and a combination of uni-gram and bi-gram. Using a composite feature created using uni-gram and bi-gram models is computationally expensive due to increased feature vector length. For this reason and the capabilities of our laptop, we used only for this stage a balanced sample of the dataset with 1200 reviews. Fourth, the effect of every individual feature selection method was evaluated based on different thresholds. The thresholds were set based on Top-K features, where the range of K values is between 1000 and 4500. Fifth, we selected two feature selection methods with the highest results, and combined them in a sequence mode. The key idea of the combination in this mode is to exploit their advantages and avoid their disadvantages.

### 4.2 Evaluation

In order to evaluate the performance, the N-fold cross validation was employed. Taking into account our device's capabilities, we used 5-fold cross validation to evaluate the performances of the first four stages. The whole dataset was divided into five sets with equal sized samples, where the classifier was trained on four sets and the remaining set was used for testing. Regarding the fifth stage, a 10-fold cross validation was used to evaluate the performance, as by this stage the computational difficulty has been already reduced by selecting the most relevant subset of features and the best text representation. To measure the performance of the SVM classifier, the following evaluation metrics were chosen: Accuracy, Precision, and Recall for evaluating the SVM classifier; see Equations 10, 11 and 12.

$$\text{Accuracy} = \frac{TP + TN}{TP + FP + TN + FN} \quad (10)$$

$$\text{Precision} = \frac{TP}{TP + FP} \quad (11)$$

$$\text{Recall} = \frac{TP}{TP + FN} \quad (12)$$

Where TP indicates a true positive which means the number of the inputs in data test that have been classified as positive when they are really belong to the positive class. TN indicates a true negative which means the number of the inputs in data test that have been classified as negative when they are really belong to the negative class. FP indicate a false positive which means the number of the inputs in data test that have been classified as positive when they are really belong to the negative class. FN indicates a false negative which means the number of the inputs in data test that have been classified as negative when they are really belong to the positive class.

### 4.3 Results

In this section, we report the experimental results of assessing the performance of the classifier. The goal is to investigate and compare different structures and representations when an SVM classifier is used. Table 1 shows the results of investigating the three weighting schemes TF-IDF, TF, and BTO when a 5-Fold cross validation is used. The results of the comparison showed that the classifier performed better with TF-IDF in terms of accuracy and recall compared to TF and BTO. It can be seen that the precision based on BTO was the highest. That means a high number of positive documents were correctly classified as positive. However, the results shown in the table are not decisive since it may be different to other datasets. It should be mentioned that no feature selection methods were used in this experiment, and the created features were 12261. For the following experiments, the TF-IDF with accuracy 87.63% was selected as a vector representation.

Table 1: Uni-gram weighting schemes cross validation results.

| Weighting Schemes | Accuracy (%) | Precision (%) | Recall (%) |
|---|---|---|---|
| TF-IDF | 87.63 | 88.10 | 87 |
| TF | 86.83 | 87.79 | 85.58 |
| BTO | 84.21 | 90.04 | 77.00 |

Table 2 shows the results after applying some pre-processing techniques such as stop word removal, light stemming, and root stemming. The results showed that the highest performance was obtained only after removing the stop words with accuracy 87.96%. Although, using the two different stemmers with stop words removal reduced the dimensionality and the noise in the data, the performance of the classifier decreased. This can be interpreted by the fact that these stemmers meant to be for MSA, and cannot be applied to Arabic dialects. For the following experiments, only stop words removal was considered in the classification process.

Table 2: The cross validation results after Applying different pre-processing techniques to uni-gram features.

| Pre-Process | Feature No. | Accuracy (%) | Precision (%) | Recall (%) |
|---|---|---|---|---|
| Stop words removal | 12072 | 87.96 | 88.43 | 87.33 |
| Stop words removal+ Light stemming | 8650 | 87.71 | 88.26 | 87 |
| Stop words+ Root stemming | 3342 | 86.58 | 87.20 | 85.75 |

Table 3 shows the results of experimenting with uni-gram, bi-gram, and a combination of both. Due to memory limitations, we randomly selected from the dataset a balanced sample set consists of 1200 reviews to assess the N-gram models. It can been seen that using uni-gram or the combination of uni-gram and bi-gram produced the same accuracy results except a slight difference in precision in favor of uni-gram and slight difference in recall in favor of the combination. That means that the classifier did not learn much from the combination of the uni-gram and bi-gram models. Using only bi-gram model displayed lower results with accuracy 71.33%, which can means some nonessential relationships were created between the words. For the following experiments, we considered the uni-gram with accuracy 85.17 as the best choice to be the feature vector model.

Table 3: The cross validation results of n-gram models.

| N-gram Model | Feature No. | Accuracy (%) | Precision (%) | Recall (%) |
|---|---|---|---|---|
| Uni-gram | 7648 | 85.17 | 86.63 | 83.33 |
| Bi-gram | 19074 | 71.33 | 72.50 | 74.17 |
| Uni-gram+Bi-gram | 26722 | 85.17 | 86.41 | 83.50 |

Table 4 shows the results of employing five feature selection methods when the SVM is used. Each feature selection method was experimented with eight subsets with

sizes of 1000, 1500, 2000, 2500, 3000, 3500, 4000, and 4500. The results indicated that feature selection methods IG and Gini index behaved in a similar mode, where the larger the number of features the lower the performance. In Figure 1, we can see that IG and Gini index obtained the highest performance when the number of features was 1500 which is less than the number of instances with accuracy over 90% for the both. When Chi-Square was used, the SVM classifier showed the lowest results compared to the other four feature selection methods. The results of Chi-Square indicated that the performance increases while the number of features increases until a specific point then start falling down. With Chi-Square, the highest accuracy of the classifier was 88.63%, when the features subset was set to 4000 which are higher than the number of instances. It can be seen that the second highest performance of the classifier was obtained after the correlation had been used. The highest accuracy of the classifier was 91.29%, when the top correlated feature subset was set to 3500 which is greater than the number of instances. Finally, it can be noticed that after experimenting the SVM with all five feature selection methods, the SVM feature selection method gave the best accuracy, precision, and recall, as depicted in Figure 1. The highest accuracy recorded was 92.12% when the feature subset is set to 4000 which is greater than the number of instances. The high results obtained after using SVM feature selection methods along with the SVM classifier can be interpreted as they use the same prediction model. Based on these findings, it can be noticed that the feature selection methods SVM and correlation gave best results of accuracy, precision, and recall among the five investigated methods when the SVM classifier is used. Therefore, we continued with the last experiment in which the SVM and correlation selection methods combined in a sequence mode to investigate the performance of the SVM classifier.

Table 4: The cross validation results after applying different pre-processing techniques to uni-gram features.

| FSM | Top-K | Accuracy (%) | Precision (%) | Recall (%) |
|---|---|---|---|---|
| IG | 1000 | 89.38 | 90.85 | 87.58 |
|  | 1500 | 90.42 | 91.47 | 89.17 |
|  | 2000 | 89.96 | 91.10 | 88.58 |
|  | 2500 | 90.12 | 90.98 | 89.08 |
|  | 3000 | 90.25 | 91.22 | 89.08 |
|  | 3500 | 90 | 90.08 | 89.92 |
|  | 4000 | 89.92 | 90.12 | 89.67 |
|  | 4500 | 90.13 | 90.30 | 89.92 |
| Correlation | 1000 | 89.83 | 92.07 | 87.17 |
|  | 1500 | 90.50 | 92.56 | 88.08 |
|  | 2000 | 90.96 | 92.85 | 88.75 |
|  | 2500 | 91.17 | 93.03 | 89.00 |
|  | 3000 | 91.08 | 92.58 | 89.33 |
|  | 3500 | 91.29 | 92.68 | 89.67 |
|  | 4000 | 91.13 | 92.43 | 89.58 |
|  | 4500 | 91.17 | 92.51 | 89.58 |
| Chi-Square | 1000 | 86.50 | 87.44 | 85.25 |
|  | 1500 | 87.58 | 87.86 | 87.25 |
|  | 2000 | 87.79 | 87.81 | 87.75 |
|  | 2500 | 88.00 | 88.51 | 87.33 |
|  | 3000 | 87.87 | 88.22 | 87.42 |
|  | 3500 | 88.17 | 88.41 | 87.83 |
|  | 4000 | 88.63 | 88.78 | 88.42 |
|  | 4500 | 87.58 | 87.70 | 87.42 |
| SVM | 1000 | 91.58 | 92.87 | 90.08 |
|  | 1500 | 91.71 | 93.27 | 89.92 |
|  | 2000 | 91.54 | 93.10 | 89.75 |
|  | 2500 | 91.67 | 93.18 | 89.92 |
|  | 3000 | 91.88 | 92.92 | 90.67 |
|  | 3500 | 91.67 | 92.81 | 90.33 |
|  | 4000 | 92.12 | 92.96 | 91.17 |
|  | 4500 | 91.83 | 92.69 | 90.83 |
| Gini Index | 1000 | 89.04 | 90.50 | 87.25 |
|  | 1500 | 90.29 | 90.88 | 89.58 |
|  | 2000 | 89.88 | 91.02 | 88.50 |
|  | 2500 | 89.92 | 90.79 | 88.83 |
|  | 3000 | 90.08 | 90.76 | 89.25 |
|  | 3500 | 90.00 | 90.08 | 89.92 |
|  | 4000 | 89.17 | 89.42 | 88.83 |
|  | 4500 | 89.58 | 89.71 | 89.42 |

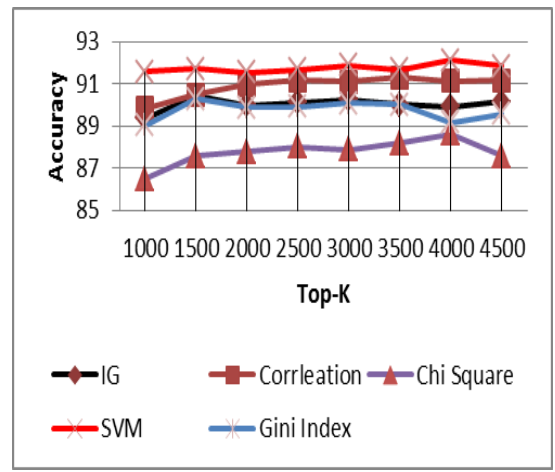

Fig. 1 Performance of the SVM classifier with the feature selection methods.

Figure 2 illustrates the performance of the SVM classifier when the two combinations of feature selection methods (SVM and correlation) were used. More details about results were presented in Table 5 and 6. Table 5 shows the results of the classifier when the correlation selection method was the first in the sequence followed by the SVM selection method. Therefore, the number of top selected features for the correlation method was set to 3500, as it is the number with which the classifier had obtained the highest performance in the previous experiment. The resulting features subset from correlation method served as the main features set from which the SVM method will select the top relevant features. The top selected features for SVM varied from 1000 to 3500 to find the subset with which the performance may improve. As shown in Table 5, the highest performance obtained when the SVM selection method was set to 1500 with accuracy 93.25%.

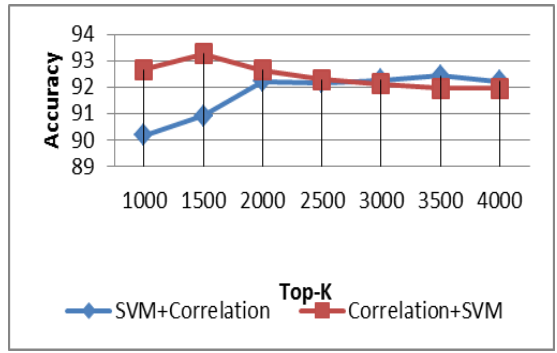

Figure 2: Performance of the SVM classifier with the combined feature selection methods.

Table 6 presents the results of another experiment that investigated the combination of the feature selection methods SVM and correlation. But In this experiment the SVM selection method was the first in the sequence followed by the SVM selection method. Here, the number of top selected features for the SVM method was set to 4000, as it is the number with which the classifier had obtained the highest performance in the previous experiment. The top selected features for correlation varied from 1000 to 4000 to find the subset with which the performance may improve. As shown in Table 6, the highest performance obtained when the correlation selection method was set to 1500 with accuracy 92.42%.

Table 5: The results of combining feature selection methods (Correlation + SVM) when the top-k features for correlation is set to 3500.

|   | Top-K Features (SVM) | Accuracy (%) | Precision (%) | Recall (%) |
|---|---|---|---|---|
| Correlation + SVM | 1000 | 92.67 | 94.21 | 91.00 |
|   | 1500 | 93.25 | 94.48 | 91.92 |
|   | 2000 | 92.63 | 94.17 | 90.92 |
|   | 2500 | 92.29 | 93.80 | 90.58 |
|   | 3000 | 92.12 | 93.50 | 90.58 |
|   | 3500 | 91.96 | 93.33 | 90.42 |

Table 6: The results of combining feature selection methods (SVM + Correlation) when the top-k features for SVM is set to 4000.

|   | Top-K Features (Correlation) | Accuracy (%) | Precision (%) | Recall (%) |
|---|---|---|---|---|
| SVM + Correlation | 1000 | 90.17 | 91.86 | 88.25 |
|   | 1500 | 90.92 | 92.64 | 88.92 |
|   | 2000 | 92.21 | 93.64 | 90.58 |
|   | 2500 | 92.17 | 93.58 | 90.58 |
|   | 3000 | 92.25 | 93.89 | 90.42 |
|   | 3500 | 92.42 | 93.99 | 90.67 |
|   | 4000 | 92.21 | 93.38 | 90.92 |

## 3. Conclusion

This work investigated the effect of five feature selection methods on the performance of the SVM classifier for dialectal Arabic sentiment analysis. As well, we combined some of the selection methods to explore their ability to improve the features selection. Additionally, the effect of different term weighting schemes, stemmers, stop words removal, and feature models on the performance were investigated. The results of investigating the individual selection methods show that the SVM classifier obtained the highest performance when the SVM selection method was used. The results also show that combining the feature selection method can improve the classifier performance. In this work, the SVM classifier obtained the highest performance when the correlation and SVM selection methods were combined in a sequence mode. Also it can be seen from the results that the SVM classifier obtained high results with a uni-gram model, TF-IDF, and stop word removal.

## References

[1] Liu, B., "Sentiment Analysis and Subjectivity". Handbook of natural language processing, 2010. 2: p. 627-666.
[2] Al-Sughaiyer, I.A. and I.A. Al-Kharashi, "Arabic morphological analysis techniques: A comprehensive survey". Journal of the American Society for Information Science and Technology, 2004. 55(3): p. 189-213.
[3] El-Beltagy, S.R. and A. Ali, "Open issues in the sentiment analysis of Arabic social media: A case study". in 2013 9th International Conference on Innovations in Information Technology (IIT). 2013. IEEE.
[4] Harrington, P., "Machine learning in action". Shelter Island, NY: Manning Publications Co, 2012.
[5] Pang, B., L. Lee, and S. Vaithyanathan, "Thumbs up?: sentiment classification using machine learning techniques". in Proceedings of the ACL-02 conference on Empirical

methods in natural language processing-Volume 10. 2002. Association for Computational Linguistics.
[6] Agarwal, B. and N. Mittal, "Prominent feature extraction for review analysis: an empirical study". Journal of Experimental & Theoretical Artificial Intelligence, 2016. 28(3): p. 485-498.
[7] Abbasi, A., H. Chen, and A. Salem, "Sentiment analysis in multiple languages: Feature selection for opinion classification in Web forums". ACM Trans. Inform. Syst, 2008. 26(3): p. 12.
[8] Ibrahim, H.S., S.M. Abdou, and M. Gheith, "Sentiment analysis for modern standard Arabic and colloquial". International Journal on Natural Language Computing (IJNLC), 2015. 4(2): p. 95-109.
[9] Shoukry, A. and A. Rafea, "A hybrid approach for sentiment classification of Egyptian Dialect Tweets". in 2015 First International Conference on Arabic Computational Linguistics (ACLing). 2015. IEEE.
[10] Aldayel, H.K. and A. Azmi, "Arabic tweets sentiment analysis–a hybrid scheme". Journal of Information Science, 2016. 42(6): p. 782-797.
[11] Al-Harbi, O., "Using objective words in the reviews to improve the colloquial arabic sentiment analysis". International Journal on Natural Language Computing (IJNLC), 2017. 6(3): p. 1-14.
[12] Guyon, I. and A. Elisseeff, "An introduction to variable and feature selection". Journal of machine learning research, 2003. 3(Mar): p. 1157-1182.
[13] Gamon, M., "Sentiment classification on customer feedback data: noisy data, large feature vectors, and the role of linguistic analysis". in Proceedings of the 20th international conference on Computational Linguistics. 2004. Association for Computational Linguistics.
[14] Kennedy, A. and D. Inkpen, "Sentiment classification of movie reviews using contextual valence shifters". Computational intelligence, 2006. 22(2): p. 110-125.
[15] Agarwal, A., et al., "Sentiment analysis of twitter data". in Proceedings of the workshop on languages in social media. 2011. Association for Computational Linguistics.
[16] Forman, G., "An extensive empirical study of feature selection metrics for text classification". Journal of machine learning research, 2003. 3(Mar): p. 1289-1305.
[17] Yu, L. and H. Liu, "Efficient feature selection via analysis of relevance and redundancy". Journal of machine learning research, 2004. 5(Oct): p. 1205-1224.
[18] Agarwal, B. and N. Mittal, "Sentiment classification using rough set based hybrid feature selection". in Proceedings of the 4th Workshop on Computational Approaches to Subjectivity, Sentiment and Social Media Analysis. 2013.
[19] Tang, J., S. Alelyani, and H. Liu, "Feature selection for classification: A review". Data classification: Algorithms and applications, 2014: p. 37.
[20] Liu, H. and H. Motoda, Feature selection for knowledge discovery and data mining. Vol. 454. 2012: Springer Science & Business Media.
[21] Pang, B. and L. Lee, "Opinion mining and sentiment analysis". Foundations and Trends® in Information Retrieval, 2008. 2(1–2): p. 1-135.
[22] Lin, Y., et al., "An information theoretic approach to sentiment polarity classification". in Proceedings of the 2nd joint WICOW/AIRWeb workshop on web quality. 2012. ACM.
[23] Moraes, R., J.F. Valiati, and W.P.G. Neto, "Document-level sentiment classification: An empirical comparison between SVM and ANN". Expert Systems with Applications, 2013. 40(2): p. 621-633.
[24] Deng, Z.-H., K.-H. Luo, and H.-L. Yu, "A study of supervised term weighting scheme for sentiment analysis". Expert Systems with Applications, 2014. 41(7): p. 3506-3513.
[25] O'Keefe, T. and I. Koprinska, "Feature selection and weighting methods in sentiment analysis". in Proceedings of the 14th Australasian document computing symposium, Sydney. 2009. Citeseer.
[26] Wang, S., et al., "A feature selection method based on fisher's discriminant ratio for text sentiment classification". in International Conference on Web Information Systems and Mining. 2009. Springer.
[27] Nicholls, C. and F. Song, "Comparison of feature selection methods for sentiment analysis". in Canadian Conference on Artificial Intelligence. 2010. Springer.
[28] Agarwal, B. and N. Mittal, "Categorical probability proportion difference (CPPD): a feature selection method for sentiment classification". in Proceedings of the 2nd workshop on sentiment analysis where AI meets psychology, COLING. 2012.
[29] Duwairi, R. and M. El-Orfali, "A study of the effects of preprocessing strategies on sentiment analysis for Arabic text". Journal of Information Science, 2014. 40(4): p. 501-513.
[30] Khalil, T., et al., "Which configuration works best? an experimental study on supervised Arabic twitter sentiment analysis". in 2015 First International Conference on Arabic Computational Linguistics (ACLing). 2015. IEEE.
[31] Aliane, A., et al., "A genetic algorithm feature selection based approach for Arabic Sentiment Classification". in IEEE/ACS 13th International Conference of Computer Systems and Applications (AICCSA). 2016. IEEE.
[32] El-Naggar, N., Y. El-Sonbaty, and M.A. El-Nasr, "Sentiment analysis of modern standard Arabic and Egyptian dialectal Arabic tweets". in 2017 Computing Conference. 2017. IEEE.
[33] Mourad, A. and K. Darwish, "Subjectivity and sentiment analysis of modern standard Arabic and Arabic microblogs". in Proceedings of the 4th workshop on computational approaches to subjectivity, sentiment and social media analysis. 2013.
[34] El-Makky, N., et al., "Sentiment analysis of colloquial Arabic tweets". in ASE BigData/SocialInformatics/PASSAT/BioMedCom 2014 Conference, Harvard University. 2014.
[35] Omar, N., et al., "A comparative study of feature selection and machine learning algorithms for arabic sentiment classification". in Asia information retrieval symposium. 2014. Springer.
[36] Mierswa, I., et al., "Yale: Rapid prototyping for complex data mining tasks". in Proceedings of the 12th ACM SIGKDD international conference on Knowledge discovery and data mining. 2006. ACM.

[37] Pak, A. and P. Paroubek, "Twitter as a corpus for sentiment analysis and opinion mining". in LREc. 2010.
[38] Shang, W., et al., "A novel feature selection algorithm for text categorization". Expert Systems with Applications, 2007. 33(1): p. 1-5.
[39] Haralampieva, V. and G. Brown, "Evaluation of Mutual information versus Gini index for stable feature selection". 2016.
[40] Bron, E.E., et al., "Feature Selection Based on the SVM Weight Vector for Classification of Dementia". IEEE J. Biomed. Health Informat, 2015. 19(5): p. 1617-1626.
[41] Pang, B. and L. Lee, "A sentimental education: Sentiment analysis using subjectivity summarization based on minimum cuts". in Proceedings of the 42nd annual meeting on Association for Computational Linguistics. 2004. Association for Computational Linguistics.
[42] Cortes, C. and V. Vapnik, "Support-vector networks". Machine Learning, 1995. 20(3): p. 273-297.
[43] Joachims, T., "Text categorization with support vector machines: Learning with many relevant features". in European conference on machine learning. 1998. Springer.
[44] Chang, C.-C. and C.-J. Lin. LIBSVM: a library for support vector machines. 2001 [cited 2 3]; 27]. Available from: www.csie.ntu.edu.tw/~cjlin/libsvm/faq.html.

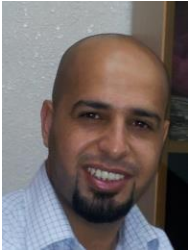
**Omar Al-Harbi** received the Ph.D. degree in Computer Science from Islamic Science University of Malaysia (USIM) in 2013. During 1997-1999, he is currently an assistant professor at the Department of Computer Science at Community College/Jazan University, Saudi Arabia Kingdom. His current research interests include word sense disambiguation, ontology, information retrieval, sentiment analysis, and question answering systems.